\documentclass{article}
\usepackage{ijcai22}

\usepackage{times}
\usepackage{soul}
\usepackage{url}
\usepackage[hidelinks]{hyperref}
\usepackage[utf8]{inputenc}
\usepackage[small]{caption}
\usepackage{graphicx}
\usepackage{amsmath}
\usepackage{amsthm}
\usepackage{booktabs}
\usepackage{algorithm}
\usepackage{algorithmic}
\usepackage{xcolor}

\newcommand{\nosection}[1]{\vspace{2pt}\noindent\textbf{#1.}}
\newcommand{\modelname}{VFGNN}
\newcommand{\modelnames}{VFGNNs}
\newcommand{\myfont}{\fontsize{8.8pt}{\baselineskip}\selectfont}
\usepackage{enumitem}

\usepackage{booktabs}
\usepackage{subfigure}
\usepackage{dsfont}
\usepackage{amssymb} 
\usepackage{amsfonts}

\newtheorem{theorem}{Theorem}
\newtheorem{definition}{Definition}
\newcommand{\D}{\mathcal{D}}
\newcommand{\R}{\mathbb{R}}
\newcommand{\M}{\mathcal{M}}
\newcommand{\GN}{\mathcal{N}}

\title{Vertically Federated Graph Neural Network for Privacy-Preserving Node Classification}

\author{
Chaochao Chen$^{1}$\and
Jun Zhou$^{1,2}$\and
Longfei Zheng$^2$\and
Huiwen Wu$^2$\and
Lingjuan Lyu$^3$\and\\
Jia Wu$^4$\and
Bingzhe Wu$^5$\and
Ziqi Liu$^2$\and
Li Wang$^2$\and
Xiaolin Zheng$^{1,6}$\footnote{Corresponding Author}
\affiliations
$^1$Zhejiang University, 
$^2$Ant Group, 
$^3$Sony AI,
$^4$Macquarie University,\\
$^5$Peking University,
$^6$JZTData Technology
\emails
\{zjuccc, xlzheng\}@zju.edu.cn,
Lingjuan.Lv@sony.com,
jia.wu@mq.edu.au, wubingzhe@pku.edu.cn, \\ \{jun.zhoujun, zlf206411, huiwen.whw, ziqiliu, raymond.wangl\}@antgroup.com 
}

\begin{document}

\maketitle

\begin{abstract}
Graph Neural Network~(GNN) has achieved remarkable progresses in various real-world tasks on graph data. 
High-performance GNN models always depend on both rich features and complete edge information in graph.  
However, such information could possibly be isolated by different data holders in practice, which is the so-called data isolation problem. 
To solve this problem, in this paper, we propose Vertically Federated Graph Neural Network (\modelname), a federated GNN learning paradigm for privacy-preserving node classification task under data vertically partitioned setting, which can be generalized to existing GNN models. 
Specifically, we split the computation graph into two parts. 
We leave the private data (i.e., features, edges, and labels) related computations on data holders, and delegate the rest of computations to a semi-honest server. 
We also propose to apply differential privacy to prevent potential information leakage from the server. 
We conduct experiments on three benchmarks and the results demonstrate the effectiveness of \modelname. 
\end{abstract}

\section{Introduction}\label{sec-intro}

Graph Neural Network (GNN) has gained increasing attentions from both academy and industry due to its ability to model high-dimensional feature information and high-order adjacent information on both homogeneous and heterogeneous graphs \cite{wu2019comprehensive}. 
%
%
An important ingredient for high-performance GNN models is high-quality graph data including rich node features and complete adjacent information.  However, in practice, such information could possibly be isolated by different data holders, which is the so-called \emph{data isolation} problem. 
Such a data isolation problem presents a serious challenge for the development of Artificial Intelligence, which becomes a hot research topic recently.

\begin{figure}
\centering
\includegraphics[width=8cm]{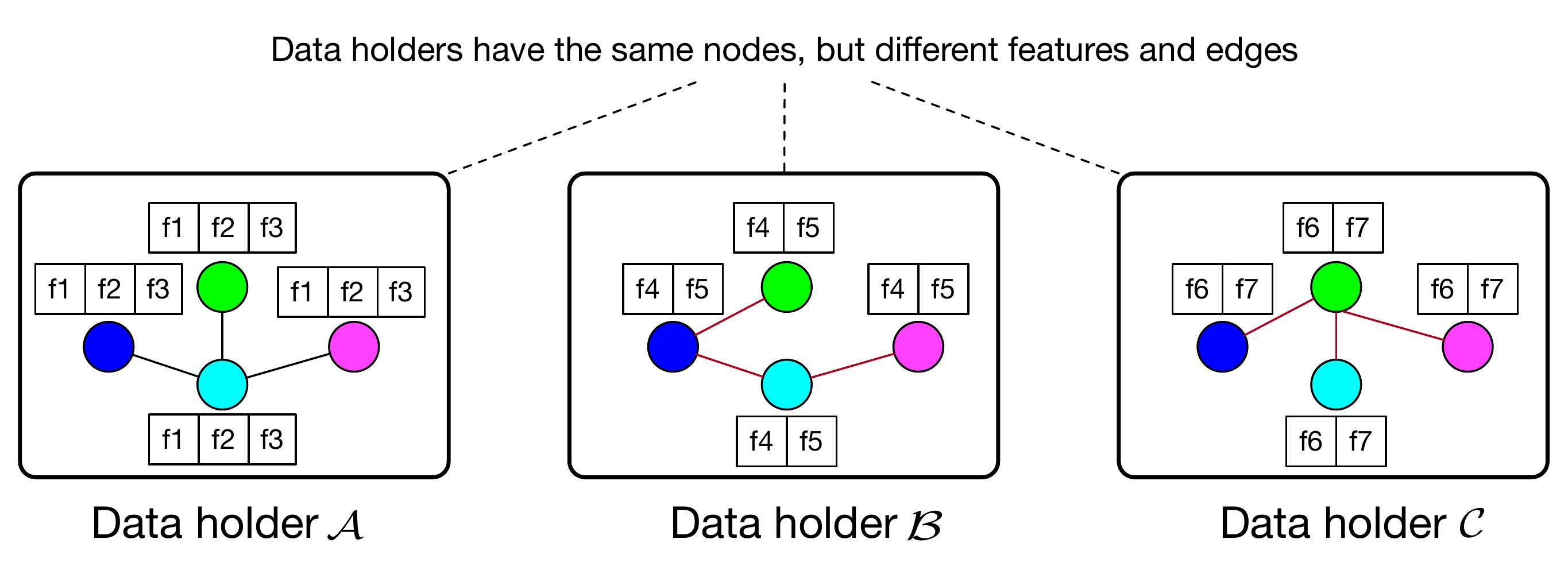}
\vskip -0.13in
\caption{Problem description of vertically federated GNN.}
\label{example}
\vskip -0.15in
\end{figure}

\nosection{Problem} 
Figure \ref{example} shows a privacy-preserving node classification problem under vertically partitioned data setting. 
Here, we assume there are three data holders ($\mathcal{A}$, $\mathcal{B}$, and $\mathcal{C}$) and they have four same nodes. The node features are vertically split, i.e., $\mathcal{A}$ has $f1$, $f2$, and $f3$, $\mathcal{B}$ has $f4$ and $f5$, and $\mathcal{C}$ has $f6$ and $f7$. Meanwhile, $\mathcal{A}$, $\mathcal{B}$, and $\mathcal{C}$ may have their own edges. 
For example, $\mathcal{A}$ has social relation between nodes while $\mathcal{B}$ and $\mathcal{C}$ have payment relation between nodes. 
We also assume $\mathcal{A}$ is the party who has the node labels. 
%
The problem is to build federated GNN models using the graph data of $\mathcal{A}$, $\mathcal{B}$, and $\mathcal{C}$. 

\nosection{Related work} 
To date, many kinds of privacy-preserving machine learning models have been proposed, e.g., logistic regression \cite{chen2021when}, decision tree \cite{fang2021large}, 
and neural network \cite{wagh2019securenn}. 
There are also several work on studying the privacy issues in GNN, e.g., graph publishing \cite{sajadmanesh2020locally}, GNN inference \cite{he2020stealing}, and federated GNN when data are horizontally partitioned \cite{zheng2021asfgnn,wu2021fedgnn}. 
%
%
So far, few research has studied the problem of GNN when data are vertically partitioned, which popularly exists in practice. 
Unlike previous privacy-preserving machine learning models that assume only samples (nodes) are held by different parties and these samples have no relationship, our task is more challenging because GNN relies on the relationships between samples, which are also kept by different data holders.

\nosection{Naive solution} 
A direct way to build privacy-preserving GNN is employing advanced cryptographic algorithms, such as homomorphic encryption~(HE) and secure multi party computation~(MPC) \cite{mohassel2017secureml}. 
Such a pure cryptographic way can provide high security guarantees, however, it suffers high computation and communication costs, which limits their efficiency  \cite{osia2019hybrid}. 

\nosection{Our solution} 
Instead, we propose \modelname, a federated GNN learning paradigm for privacy-preserving node classification task under data vertically partitioned setting. 
Motivated by the existing work in split learning \cite{vepakomma2018split,osia2019hybrid,gu2018securing}, we split the computation graph of GNN into two parts for privacy and efficiency concern, i.e., the private data related computations carried out by data holders and non-private data related computations conducted by a semi-honest server. 

Specifically, data holders first apply MPC techniques to compute the initial layer of the GNN using private node feature information collaboratively, which acts as the feature extractor module, and then perform neighborhood  aggregation using private edge information individually, similar as the existing GNNs \cite{velivckovic2017graph}, and finally get the local node embeddings. 
Next, we propose different combination strategies for a semi-honest server to combine local node embeddings from data holders and generate global node embeddings, based on which the server can conduct the successive non-private data related computations, e.g., the non-linear operations in deep network structures that are time-consuming for MPC techniques. 
Finally, the server returns the final hidden layer to the party who has labels to compute prediction and loss. 
Data holders and the server perform forward and back propagations to complete model training and prediction, during which the private data (i.e., features, edges, and labels) are always kept by data holders themselves. 
Here we assume data holders are honest-but-curious, and the server does not collude with data holders. 
We argue that this is a reasonable assumption since the server can be played by authorities such as governments or replaced by trusted execution environment \cite{costan2016intel}. 
Moreover, we adopt differential privacy, on the exchanged information between server and data holders (e.g., local node embeddings and gradient update), 
to further protect potential information leakage from the server.


\nosection{Contributions} 
We summarize our main contributions as: 
\begin{itemize}[leftmargin=*] \setlength{\itemsep}{-\itemsep}
\item We propose a novel \modelname~learning paradigm, which not only can be generalized to most existing GNNs, but also enjoys good accuracy and efficiency. 
\item We propose different combination strategies for the server to combine local node embeddings from data holders.
\item We evaluate our proposals on three real-world datasets, and the results demonstrate the effectiveness of \modelname. 
\end{itemize}


\section{Preliminaries}\label{preli}

\subsection{Security Model}
In this paper, we assume the adversary is honest-but-curious (semi-honest). 
That is, data holders and the server strictly follow the protocol, but they also use all intermediate computation results to infer as much information as possible. We also assume that the server does not collude with any data holders. This security setting is similar as most existing work \cite{mohassel2017secureml,hardy2017private}. 


%
%
%

\subsection{Additive Secret Sharing}\label{pre-ss}
Additive Secret Sharing has two main procedures.   \cite{shamir1979share}. 
To additively share $\textbf{Shr}(\cdot)$ an $\ell$-bit value $a$ for party $i \in \mathcal{P} = \{1,...,I\}$, party $i$ generates \{$a_j \in \mathds{Z}_{2^\ell}, j \in \mathcal{P}$ and $j \neq i$\} uniformly at random, sends $a_j$ to party $j$, and keeps $a_i = a- \sum_j a_j$ mod $2^\ell$. We use $\langle a \rangle _i = a_i$ to denote the share of party $i$. To reconstruct $\textbf{Rec}(\cdot, \cdot)$ a shared value $\langle a \rangle$, each party $i$ sends $\langle a \rangle _i$ to one who computes $\sum_i a_i$ mod $2^\ell$. 
For simplification, we denote additive sharing by $\langle \cdot \rangle$. 
Addition in secret sharing can be done by participants locally. 
Multiplication in secret sharing usually relies on Beaver's triplet technique \cite{beaver1991efficient}.

\subsection{Differential Privacy}\label{sec:pre_dp}


\begin{definition}\label{def:dp} (Differential Privacy \cite{dwork2014algorithmic}). 
A randomized algorithm $\M$ that takes as input a dataset consisting of individuals is $(\epsilon, \delta)-$differentially private (DP) if for any pair of neighbouring data $x, y$ that differ in a single entry, and any event $E$, 
\begin{equation}\small
P [\mathcal{M}(x) \in E ] \leq \exp(\epsilon) P[\mathcal{M}(y) \in E] + \delta, 
\end{equation}
and if $\delta = 0$, we say that $\mathcal{M}$ is $\epsilon-$differentially private. 
\end{definition}





In \cite{dwork2014algorithmic}, the authors pointed out that the $\ell_2-$sensitivity of a function $f$ measures the magnitude by which a single individual's data can change the function in the worst case. 
%
%

\begin{definition} \label{def:l2_sens} ($\ell_2$-sensitivity ~\cite{dwork2014algorithmic}). 
Suppose $x$ and $y$ are neighbouring inputs that differ in one entry. 
The $\ell_2$-sensitivity of  a function $f: \D \rightarrow \R^d$ is:
\begin{equation}\small
\Delta_2 f = \max_{x,y \in \D, \|x - y \| =1} \| f(x) - f(y)\|_2.
\end{equation}
\end{definition}

\begin{definition}\label{def:gaussian_mecha} (The Gaussian Mechanism ~\cite{dwork2014algorithmic})
Given a function $f: \D \rightarrow \R^d$ over a dataset $\D$, the Gaussian mechanism is defined as:
\begin{equation}\small
\M_G(x, f(\cdot), \epsilon) = f(x) + (Y_1, \cdots, Y_k),
\end{equation}
where $Y_i$ are $i.i.d.$ random variables drawn from $\mathcal{N}(0, \sigma^2 \Delta_2 f^2)$ and 
$\sigma = \frac{\sqrt{2 \ln(1.25/\delta)}}{\epsilon}$. 
\end{definition}

\begin{theorem}~\cite{dwork2014algorithmic}
\label{thm:privacy_one_step}
The Gaussian mechanism defined in Definition~\ref{def:gaussian_mecha} preserves $(\epsilon, \delta)-$DP for each publication step. 
\end{theorem} 
\section{The Proposed Model}\label{model}

\subsection{Overview of \modelname}\label{sec-model-ov}

\begin{figure}
\centering
\includegraphics[width=0.95\columnwidth]{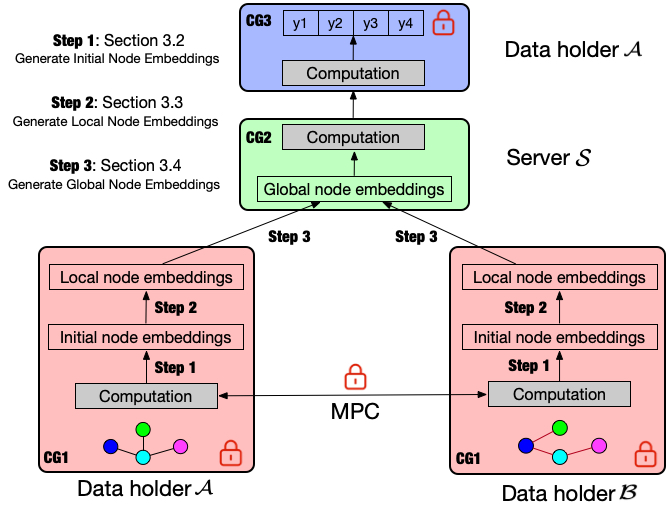}
\vskip -0.12in
\caption{Overview of our proposed \modelname.
}
\label{fig:model}
\vskip -0.15in
\end{figure}



As described in Section 1, for the sake of privacy and efficiency, we design a vertically federated GNN (\modelname) learning paradigm by \textit{splitting the computational graph of GNN} into two parts. 
That is, we keep the \textit{private data related computations} to data holders for privacy concern, and delegate the \textit{non-private data related computations} to a semi-honest server for efficiency concern. 
In the context of GNN, the private data refers to node features, labels, and edges (node relations). 
To be specific, we divide the computational graph into the following 
three sub-Computational Graphs (CG), as is shown in Figure \ref{fig:model}. 


\nosection{CG1: private feature and edge related computations} 
The first step of GNN is generating initial node embeddings using node's private features, e.g., user features in social networks. In vertically data split setting, each data holder has partial node features, as shown in Figure \ref{example}. We will present how data holders collaboratively learn initial node embeddings later. 
In the next step, data holders generate local node embeddings by aggregating multi-hop neighbors' information using different \textit{aggregator functions}.

\nosection{CG2: non-private data related computations} 
We delegate non-private data related computations to a semi-honest server for efficiency concern. 
First, the server combines the local node embeddings from data holders with different COMBINE strategies, and obtains the global node embeddings. 
Next, the server can perform the successive computations using plaintext data. 
Note that this part has many non-linear computations such as max-pooling and activation functions, which are not cryptographically friendly. 
For example, existing works approximate the non-linear activations by using either piece-wise functions that need secure comparison protocols \cite{mohassel2017secureml} or high-order polynomials \cite{hardy2017private}. Therefore, their accuracy and efficiency are limited. 
Delegating these plaintext computations to server will not only improve our model accuracy, but also significantly improve our model efficiency, as we will present in experiments. 
After this, the server gets a final hidden layer ($\textbf{z}_L$) and sends it back to the data holder who has label to calculate the prediction, where $L$ is the total number of layers of the deep neural network. 

\nosection{CG3: private label related computations} 
The data holder who has label can compute prediction using the final hidden layer it receives from the server. 
For node classification task, the Softmax activation function is used for the output layer \cite{kipf2016semi}, which is defined as
$\text{softmax}(z_c)=\frac{1}{Z}\text{exp}(z_c)$ with $c \in C$ be the node class and $Z=\sum_c \text{exp}(z_c)$.

In the following subsections, we will describe three important components of \modelname, i.e., initial node embeddings generation in \textbf{CG1}, local node embeddings generation in \textbf{CG2}, and global node embedding generation in \textbf{CG3}. 

\begin{figure}[h]
\centering
\subfigure[Individually] { \includegraphics[width=4.0cm]{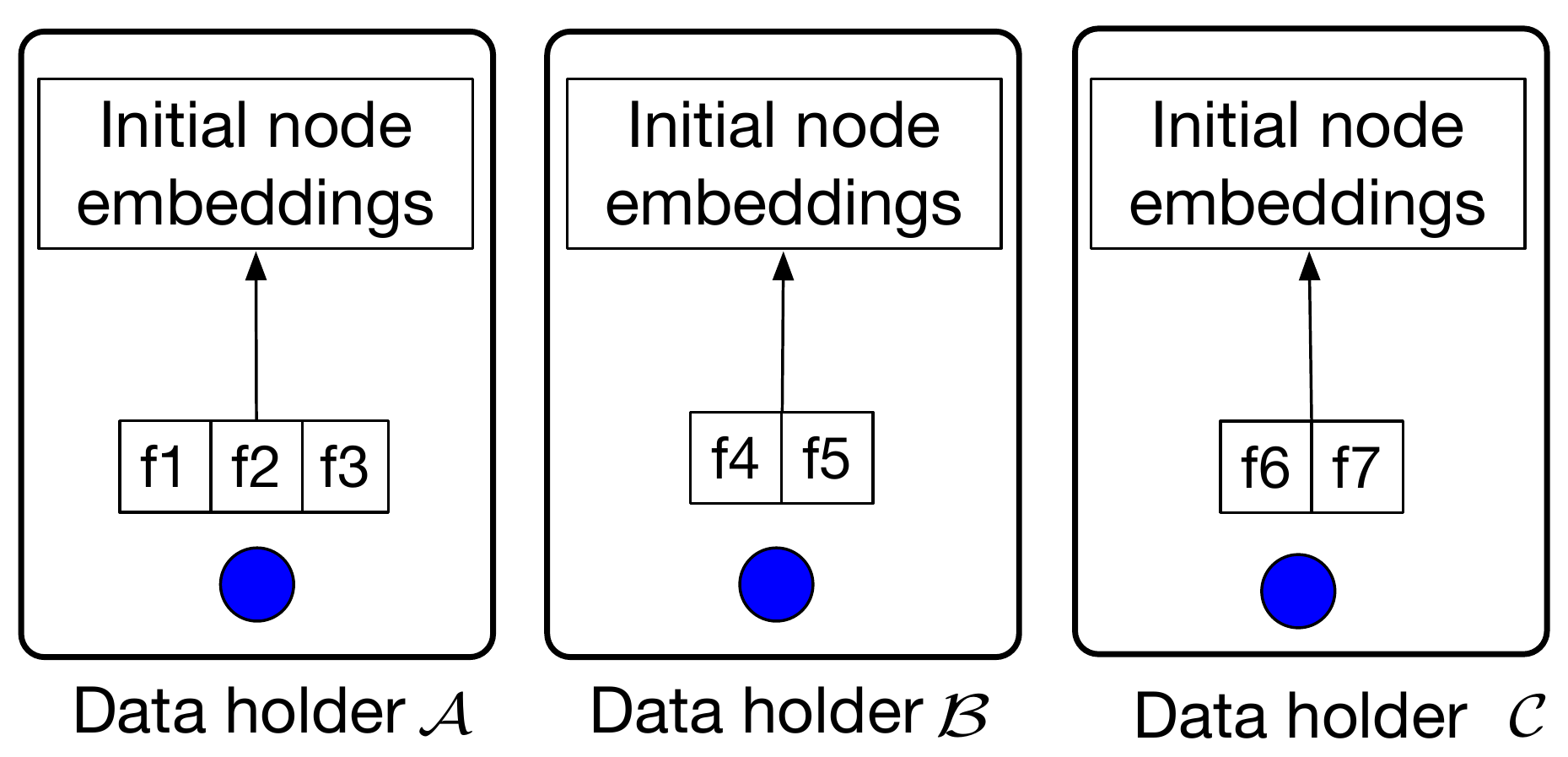}}
\subfigure[Collaboratively] { \includegraphics[width=4.0cm]{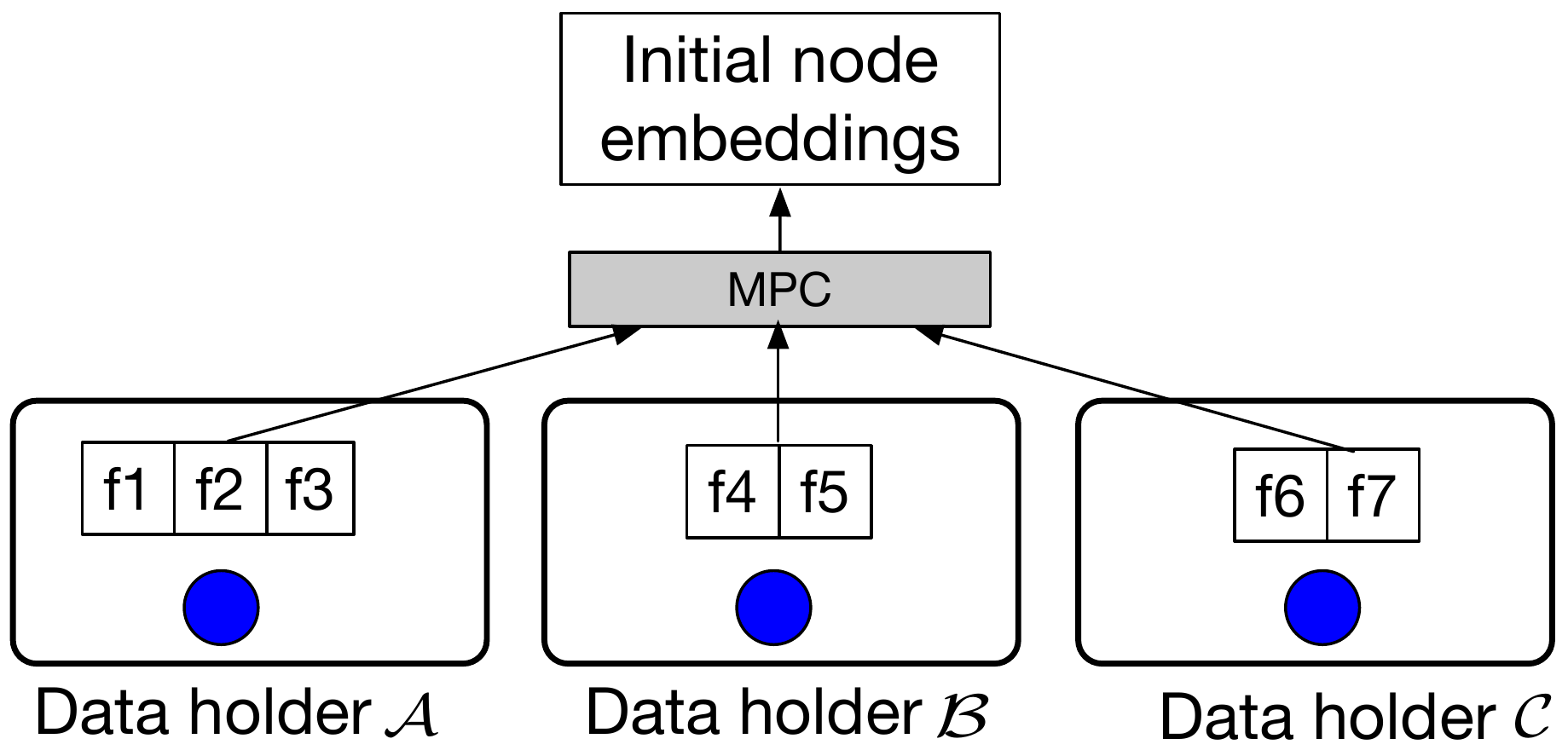}}
\vskip -0.15in
\caption{Methods of generating initial node embeddings. }
\label{fig:init-gene}
\vskip -0.12in
\end{figure}

\subsection{Generate Initial Node Embeddings}\label{sec-model-init}
Initial node embeddings are generated by using node features. 
Under vertically partitioned data setting, each data holder has partial node features. 
There are two methods for data holders to generate initial node embeddings, i.e., \textit{individually} and \textit{collaboratively}, as shown in Figure \ref{fig:init-gene}.

The `\textit{individually}' method means that data holders generate initial node embeddings using their own node features, individually.  
For data holder $i \in \mathcal{P}$, this can be done by $\textbf{h}_0^i={(\textbf{x}^i)} ^T \cdot \textbf{W}^i$, where $\textbf{x}^i$ and $\textbf{W}^i$ are node features and weight matrix of data holder $i$. 
As the example in Figure \ref{fig:init-gene} (a), $\mathcal{A}$, $\mathcal{B}$, and $\mathcal{C}$ generate their initial node embeddings using their own features separately. 
%
%
Although this method is simple and data holders do not need to communicate with each other, it cannot capture the relationship between features of data holders and thus causes information loss. 
%

To solve the shortcoming of `\textit{individually}' method, we propose a `\textit{collaboratively}' method. 
It indicates that data holders generate initial node embeddings using their node features, collaboratively, and meanwhile keep their private features secure. 
Technically, this can be done by using cryptographic methods such as secret sharing and homomorphic encryption \cite{acar2018survey}. 
In this paper, we choose additive secret sharing due to its high efficiency. 

\subsection{Generate Local Node Embeddings}\label{sec-model-aggre}
We generate local node embeddings by using multi-hop neighborhood aggregation on graphs, based on initial node embeddings. 
Note that, 
\textit{neighborhood aggregation should be done by data holders separately}, rather than cooperatively, to protect the private edge information. 
This is because one may infer the neighboorhoold information of $v$ given the neighborhood aggregation results of $k$-hop ($\textbf{h}^k_v (i)$) and $k+1$-hop ($\textbf{h}^{k+1}_v (i)$), if neighborhood aggregation is done by data holders jointly. 
%
%
%
For $\forall v \in \mathcal{V}$ at each data holder, neighborhood aggregation is the same as the traditional GNN. 
Take GraphSAGE \cite{hamilton2017inductive} for example, it introduces aggregator functions to update hidden embeddings by sampling and aggregating features from a node's local neighborhood:
\begin{equation}\small
\begin{split}
& \textbf{h}_{\mathcal{N}(v)}^k \leftarrow \text{AGG}_k(\{\textbf{h}_u^{k-1}, \forall u \in \mathcal{N}(v)\}),\\
& \textbf{h}^k_v \leftarrow (\textbf{W}^k \cdot \text{CONCAT} ( \textbf{h}^{k-1}_v,\textbf{h}_{\mathcal{N}(v)}^k ) ),
\end{split} 
\end{equation}
where we follow the same notations as GraphSAGE, and the aggregator functions AGG are of three types, i.e., Mean, LSTM, and Pooling. 
%
%
After it, data holders send local node embeddings to a semi-honest server for combination and further non-private data related computations.

\subsection{Generate Global Node Embeddings}\label{sec-model-combine}

The server combines the local node embeddings from data holders and gets global node embeddings. 
The combination strategy (COMBINE) should be trainable and maintaining high representational capacity, and we design three of them. 

\nosection{Concat}
The concat operator can fully preserve local node embeddings learnt from different data holders. That is, Line \ref{algo-learning-se-st} in Algorithm \ref{algo} becomes 
\begin{equation}\small
\textbf{h}^K_v \leftarrow \text{CONCAT} (\textbf{h}^K_v (1),\textbf{h}^K_v (2),...,\textbf{h}^K_v (I)).
\end{equation}

\nosection{Mean}
The mean operator takes the elementwise mean of the vectors in $(\{ \textbf{h}^K_v (i), \forall i \in  \mathcal{P}\})$, assuming data holders contribute equally to the global node embeddings, i.e., 
\begin{equation}\small
\textbf{h}^K_v \leftarrow \text{MEAN} (\textbf{h}^K_v (1) \cup \textbf{h}^K_v (2) \cup...\cup \textbf{h}^K_v (I)).
\end{equation}

\nosection{Regression}
The above two strategies treat data holders equally. In reality, the local node embeddings from different data holder may contribute diversely to the global node embeddings. 
We propose a Regression strategy to handle this kind of situation. 
Let $\boldsymbol{\omega_i}$ be the weight vector of local node embeddings from data holder $i \in \mathcal{P}$, then
\begin{equation}\small
\begin{split}
&\textbf{h}^K_v \leftarrow \boldsymbol{\omega_1} \odot \textbf{h}^K_v (1) + \boldsymbol{\omega_2} \odot \textbf{h}^K_v (2) ... + \boldsymbol{\omega_I} \odot \textbf{h}^K_v (I), 
\end{split}
\end{equation}
where $\odot$ is element-wise multiplication. 

These different combination operators can utilize local node embeddings in diverse ways, and we will empirically study their effects on model performances in experiments. 

\begin{algorithm}[t]
\caption{Information publishing mechanisms of data holders to server using differential privacy }\label{mecha:gauss}
\textbf{Input}: Local information of data holders $\textbf{x}$, dimension of local information $d$, noise multiplier $\sigma$, clipping value $C$. \\
\textbf{Output}: Differentially private node embeddings.

\begin{algorithmic}[1] 
\STATE Scale local information
$\mathbf{\bar{x}} = \min(1, {C}/\| \mathbf{x} \|)\mathbf{x};$
\STATE Draw i.i.d. samples from  $\GN(0, \sigma^2 C^2)$, which forms a $d$-dimension noise vector $\mathbf{n}$; 
\STATE \# \textbf{Gaussian Mechanism} 
\STATE Add noise $\mathbf{\tilde{x}}= \mathbf{x}^K + \mathbf{n};$ 
\STATE \# \textbf{James-Stein Estimator} 
\STATE Compute James-Stein Estimator \\$\mathbf{\tilde{x}_{JS}} = \left( 1 - \frac{(d-2) \sigma^2 C^2}{\| \mathbf{\tilde{x}} \|^2} \right) \mathbf{\tilde{x}}$  

\STATE \textbf{return} $\mathbf{\tilde{x}}$ or $\mathbf{\tilde{x}_{JS}}$. 
\end{algorithmic}
\end{algorithm} 

\subsection{Enhancing Privacy by Adopting DP}\label{sec:model:dp}

Data holders directly send the \textit{local information}, e.g., local node embeddings during forward propagation and gradient update during back propagation, to the server may cause potential information leakage~\cite{lyu2020threats}, and we propose to apply differential privacy to further enhance privacy. 
In this section, we introduce two DP based data publishing mechanisms, to further enhance the privacy of our proposed \modelname. 
Such that with a single entry modification in the local information of data holders, there is a large probability that the server cannot distinguish the difference before or after the modification. 
We present the two mechanisms, i.e., Gaussian Mechanism and James-Stein Estimator, in Algorithm~\ref{mecha:gauss}. 
We have described Gaussian mechanism in Section \ref{sec:pre_dp}, we present James-Stein Estimator as follows. 

%

\begin{theorem} (James-Stein Estimator and its adaptivity~\cite{balle2018improving}).
Suppose $d$ is the dimension of local information $\mathbf{x}$. 
When $d \geq 3$, substituting $w$ in $\mathbf{\tilde{x}_{Bayes}}$ with its maximum likelihood estimate under $\mathbf{x} \sim \GN(0, w^2 I)$, $\mathbf{\tilde{x}}|\mathbf{x} \sim \GN(\mathbf{x}, \sigma^2 C^2 I)$, and $\mathbf{\tilde{x}_{Bayes}} = argmin_{\mathbf{\tilde{x}}} \|\mathbf{\tilde{x}} - \mathbf{x} \|^2$ produces James-Stein Estimator
%
$
\mathbf{\tilde{x}_{JS}} = \left( 1 - \frac{(d-2) \sigma^2 C^2 }{\| \mathbf{\tilde{x}}\|^2} \right) \mathbf{\tilde{x}}. 
$
Moreover, it has an Mean Squared Error (MSE) of 
\begin{equation}\small
E [\| \mathbf{\tilde{x}_{JS}} -\mathbf{x} \|^2] = d \sigma^2 \left(1 - \frac{(d-2)^2}{d^2} \frac{\sigma^2 C^2 }{w^2 + \sigma^2 C^2 } \right).
\end{equation}
\end{theorem}

The MSE of Gaussian Mechanism $\mathbf{\tilde{x}}$ to exact $\mathbf{x}$ is $E \|\mathbf{\tilde{x}} - \mathbf{x} \|^2 = d \sigma^2 C^2$, while 
the MSE of James-Stein Estimator is reduced with a factor of $(1 - \frac{(d-2)^2}{d^2} \frac{\sigma^2 C^2 }{w^2 + \sigma^2 C^2 })$. 
Both methods preserve $(\epsilon, \delta)$-DP while James-Stein estimator shows reductions in MSE, thus improves utility. 
%
%
By the definition of Gaussian mechanism (Definition~\ref{def:gaussian_mecha}), we have the privacy loss for both information publishing mechanisms in Algorithm~\ref{mecha:gauss}. 
By combining it with Moment Accountant (MA) \cite{abadi2016deep}, we present the overall privacy for $T$ iterations.

\begin{theorem}
\label{thm:privacy_all}
Suppose each iteration of Algorithm~\ref{mecha:gauss} is $(\epsilon, \delta)-$DP. 
There exist constants $c_1$ and $c_2$ so that given the sampling probability $q$ and the number of iterations $T$, and $\epsilon < c_1 q \sqrt{T}$, 
Algorithm~\ref{mecha:gauss} over $T$ iteration is $(\epsilon', \delta)-$DP, with $\epsilon' = c_2 q \sqrt{T} \epsilon$. 
\end{theorem}
\begin{proof}
By Definition~\ref{def:gaussian_mecha} and Theorem~\ref{thm:privacy_one_step}, to ensure one iteration $(\epsilon, \delta)-$DP, we set $\sigma = \frac{\sqrt{2 \ln (1.25/\delta)}}{\epsilon}$. 
By Theorem 1 in ~\cite{abadi2016deep}, with $\sigma = \frac{\sqrt{2 \ln (1.25/\delta)}}{\epsilon}$ and the appropriate choice of $\epsilon, q, T$, such that $\epsilon < c_1 q \sqrt{T}$, the privacy loss over $T$ iterations is 
$
\epsilon' = \frac{c_1 q \sqrt{T \log(1/\delta)}}{\sigma} = c_2 q \sqrt{T} \epsilon. 
$
\end{proof}

\begin{algorithm}[t]
\caption{Privacy-preserving GraphSAGE for node label prediction (forward propagation)}\label{algo}
\textbf{Input}: Data holder $\forall i \in \mathcal{P}$; 
Graph $\mathcal{G}(\mathcal{V},\mathcal{E}^i)$ and node features \{$\textbf{x}^i_v, \forall v \in \mathcal{V}$\}; 
depth $K$; 
aggregator functions AGG$_k, \forall k \in \{1,...,K\}$; weight matrices $\textbf{W}_i^k, \forall k \in \{1,...,K\}$; 
max layer $L$; 
weight matrices $\textbf{W}_l, \forall l \in \{0,...,L\}$; 
non-linearity $\sigma$; 
neighborhood functions $\mathcal{N}^i: v \rightarrow 2^\mathcal{V}$; 
node labels on data holder $p \in \mathcal{P}$ and $c\in C$\\
\textbf{Output}: Node label predictions \{$\hat{y}_{vc}, \forall v \in \mathcal{V}, \forall c \in C$\}

\begin{algorithmic}[1] 

\STATE \# \textbf{CG1:} private feature and edge related computations 
\STATE \textbf{Data holders}: jointly calculate initial node embeddings $\textbf{h}_v^0(i) \leftarrow \textbf{x}^i_v, \forall i \in \mathcal{P}, \forall v \in \mathcal{V}$\label{algo-learning-dh-st}
\FOR{$i \in \mathcal{P}$ \texttt{in parallel}}\label{algo-learning-dh-agg}
    \FOR{$k=1$ to $K$}
        \FOR{$v \in \mathcal{V}$}
            \STATE \textbf{Data holder}: calculates $\textbf{h}_{\mathcal{N}(v)}^k (i) \leftarrow$ AGG$_k(\{\textbf{h}_u^{k-1}(i), \forall u \in \mathcal{N}^i(v)\})$ 
        \ENDFOR
        
        \STATE \textbf{Data holder}: calculates $\textbf{h}^k_v (i) \leftarrow \sigma(\textbf{W}_i^k \cdot$ CONCAT $(\textbf{h}^{k-1}_v (i),\textbf{h}_{\mathcal{N}(v)}^k (i))$ 
    \ENDFOR
    \STATE \textbf{Data holder}: calculates local node embeddings $\textbf{h}^K_v (i) \leftarrow \textbf{h}^K_v (i)/ ||\textbf{h}^K_v (i) ||_2, \forall v \in \mathcal{V}$ and sends (publishes) it to server using differential privacy
 \label{algo-learning-dh-en}
\ENDFOR

\STATE \# \textbf{CG2:} non-private data related computations 
\FOR{$v \in \mathcal{V}$}
\label{algo-learning-se-1}
    \STATE \textbf{Server}: combines the local node embeddings from data holders $\textbf{h}^K_v = $ COMBINE $(\{ \textbf{h}^K_v (i), \forall i \in  \mathcal{P}\})$\label{algo-learning-se-st}
    \STATE \textbf{Server}: forward propagation based on the global node embeddings $\textbf{z}_L = \sigma(\textbf{W}_{L-1} \cdot \sigma(... \sigma(\textbf{W}_0 \cdot \textbf{h}^K_v)))$\label{algo-learning-se-lh}
    \STATE \textbf{Server}: sends $\textbf{z}_L$ to data holder $p$
\ENDFOR \label{algo-learning-se-en}

\STATE \# \textbf{CG3:} private label related computations 
\STATE \textbf{Data holder} $p$: makes prediction by $\hat{y}_{vc} \leftarrow \text{softmax}(\textbf{W}_L \cdot \textbf{z}_L), \forall v \in \mathcal{V}, \forall c \in C$ \label{algo-learning-dh-pre}

\end{algorithmic}
\end{algorithm}

\subsection{Putting Together} 
By combining \textbf{CG1}-\textbf{CG3} and the key components described above, we complete the forward propagation of \modelname. 
To describe the procedures in details, without loss of generality, we take GraphSAGE \cite{hamilton2017inductive} 
for example and present its forward propagation process in Algorithm \ref{algo}. 
%
%
\modelname~can be learnt by gradient descent through minimizing the cross-entropy error over all labeled training examples. 

Specifically, the model weights of \modelname~are in four parts. 
(1) The \textit{weights for the initial node embeddings}, i.e., $\langle \textbf{W} \rangle _i, \forall i \in \mathcal{P}$, that are secretly shared by data holders, 
(2) the \textit{weights for neighborhoold aggregation on graphs}, i.e., $\textbf{W}_i^k$, that are also kept by data holders, 
(3) the \textit{weights for the hidden layers of deep neural networks}, i.e., $\textbf{W}_l, 0 \le l < L$, that are hold by server, 
(4) and the \textit{weights for the output layer of deep neural networks}, i.e., $\textbf{W}_L$, that are hold by the data holder who has label. 
%
%
%
As can be seen, in \modelname, both private data and model are hold by data holders themselves, thus data privacy can be better guaranteed.

\section{Experiments}\label{experiments}
We conduct experiments to answer the following questions. 
\textbf{Q1:} whether \modelname~outperforms the GNN models that are trained on the isolated data. 
\textbf{Q2:} how does \modelname~behave comparing with the traditional insecure model trained on the plaintext mixed data. 
\textbf{Q3:} how does \modelname~perform comparing with the naive solution in Section \ref{sec-intro}. 
\textbf{Q4:} are our proposed combination strategies effective to \modelname. 
\textbf{Q5:} what is the effect of the number of data holders on \modelname. 
\textbf{Q6:} what is the effect of differential privacy on \modelname. 

\subsection{Experimental Setup}

\nosection{Datasets} 
We use four benchmark datasets, i.e., Cora, Pubmed, Citeseer \cite{sen2008collective}, and arXiv \cite{hu2020open}. 
We use exactly the same dataset partition of training, validate, and test following the prior work \cite{kipf2016semi,hu2020open}. 
Besides, in data isolated GNN setting, both node features and edges are hold by different parties. 
For all the experiments, we use five-fold cross validation and adopt average accuracy as the evaluation metric. 

\begin{table}
\centering
\small
\begin{tabular}{ccccc}
  \toprule
  Dataset & \#Node & \#Edge & \#Features & \#Classes \\
  \midrule
  Cora & 2,708 & 5,429 & 1,433 & 7 \\
  Pubmed & 19,717 & 44,338 & 500 & 3 \\
  Citeseer & 3,327 & 4,732 & 3,703 & 6 \\
  arXiv & 169,343 & 2,315,598 & 128 & 40 \\
  \bottomrule
\end{tabular}
\vskip -0.1in
\caption{Dataset statistics.}
\vskip -0.15in
\label{dataset}
\end{table}

\nosection{Comparison methods} 
We compare \modelname~with GraphSAGE models \cite{hamilton2017inductive} that are trained using isolated data and mixed plaintext data to answer \textbf{Q1} and \textbf{Q2}. 
We also compare \modelname~with the naive solution described in Section 1 to answer \textbf{Q3}. 
To answer \textbf{Q4}, we vary the proportion of the data (features and edges) hold by $\mathcal{A}$ and $\mathcal{B}$, and change \modelname~with different combination strategies. 
We vary the number of data holders in \modelname~to answer \textbf{Q5}, and vary the parameters of differential privacy to answer \textbf{Q6}. 
For all these models, we choose Mean operator as the aggregator function. 

\nosection{Parameter settings} 
For all the models, we use TanH as the active function of neighbor propagation, and Sigmoid as the active function of hidden layers. 
For the deep neural network on server, we set the dropout rate to 0.5 and network structure as ($d$, $d$, $|C|$), where $d \in \{32, 64, 128\}$ is the dimension of node embeddings and $|C|$ the nubmer of classes. 
We vary $\epsilon \in \{1, 2, 4, 8, 16, 32, 64, \infty\}$, set $\delta=1e^{-4}$ and the clip value $C=1$ to study the effects of differential privacy on our model. 
Since we have many comparison and ablation models, and they achieve the best performance with different parameters, we cannot report all the best parameters. Instead, we report the range of the best parameters. 
We vary the propagation depth $K \in \{2,3,4,5\}$, L2 regularization in $\{10^{-2}-10^{-4}\}$, and learning rate in $\{10^{-2}-10^{-3}\}$. 
We tune parameters based on the validate dataset and evaluate model performance on the test dataset. 


\subsection{Comparison Results and Analysis}
To answer \textbf{Q1}-\textbf{Q3}, we assume there are two data holders ($\mathcal{A}$ and $\mathcal{B}$) who have equal number of node features and edges, i.e., the proportion of data held by $\mathcal{A}$ and $\mathcal{B}$ is 5:5, and compare our models with GraphSAGEs that are trained on isolated data individually and on mixed plaintext data. 
We also set $\epsilon=\infty$ during comparison and will study its effects later. 
We summarize the results in Table \ref{compare}, where \modelname\_C, \modelname\_M, and \modelname\_R denote \modelname~with Concat, Mean, and Regression combination strategies. 

\nosection{Result1: answer to Q1} 
We first compare \modelnames~with the GraphSAGEs that are trained on isolated feature and edge data, i.e., GraphSAGE$_\mathcal{A}$ and GraphSAGE$_\mathcal{B}$. 
From Table \ref{compare}, we find that, \modelnames~with different combination strategies significantly outperforms GraphSAGE$_\mathcal{A}$ and GraphSAGE$_\mathcal{B}$ on all the three datasets. 
Take Citeseer for example, our \modelname\_R~improves GraphSAGE$_\mathcal{A}$ and GraphSAGE$_\mathcal{B}$ by as high as 28.10\% and 51.64\%, in terms of accuracy. 

\nosection{Analysis of Result1} 
The reason of result1 is straightforward. 
GraphSAGE$_\mathcal{A}$ and GraphSAGE$_\mathcal{B}$ can only use partial feature and edge information held by $\mathcal{A}$ and $\mathcal{B}$. 
In contrast, \modelnames~provide a solution for $\mathcal{A}$ and $\mathcal{B}$ to jointly train GNNs without compromising their own data. 
By doing this, \modelnames~can use the information from the data of both $\mathcal{A}$ and $\mathcal{B}$ simultaneously, and therefore achieve better performance. 
\begin{table}[t]
\centering
\small
\begin{tabular}{ccccc}
  \toprule
  Dataset & Cora & Pubmed & Citeseer & arXiv\\
  \midrule
  \myfont{GraphSAGE$_\mathcal{A}$} & 0.611 & 0.672 & 0.541 & 0.471 \\
  \myfont{GraphSAGE$_\mathcal{B}$} & 0.606 & 0.703 & 0.457 & 0.482  \\
  \myfont{\modelname\_C} & 0.790 & 0.774 & 0.685 & 0.513  \\
  \myfont{\modelname\_M} & 0.809 & 0.781 & 0.695 & 0.522  \\
  \myfont{\modelname\_R} & 0.802 & 0.782 & 0.693 & 0.518  \\
  \myfont{GraphSAGE$_\mathcal{A+B}$} & 0.815 & 0.791 & 0.700 & 0.529  \\
  \bottomrule
\end{tabular}
\vskip -0.1in
\caption{Comparison results on three datasets (Q1 and Q2).}
\vskip -0.1in
\label{compare}
\end{table}

\nosection{Result2: answer to Q2} 
We then compare \modelnames~with GraphSAGE that is trained on the mixed plaintext data, i.e., GraphSAGE$_\mathcal{A+B}$. 
It can be seen from Table \ref{compare} that \modelnames~have comparable performance with GraphSAGE$_\mathcal{A+B}$, e.g., 0.8090 vs. 0.8150 on Cora dataset and 0.6950 vs. 0.7001 on Citeseer dataset. 

\nosection{Analysis of Result2} 
It is easy to explain why our proposal has comparable performance with the model that are trained on the mixed plaintext data. 
First, we propose a secret sharing based protocol for $\mathcal{A}$ and $\mathcal{B}$ to generate the initial node embeddings from their node features, which are the same as those generated by using mixed plaintext features. 
Second, although $\mathcal{A}$ and $\mathcal{B}$ generate local node embeddings by using their own edge data to do neighbor aggregation separately (for security concern), we propose different combination strategies to combine their local node embeddings. 
Eventually, the edge information from both $\mathcal{A}$ and $\mathcal{B}$ is used for training the classification model. Therefore, \modelname~achieves comparable performance with GraphSAGE$_\mathcal{A+B}$. 

\nosection{Result3: answer to Q3} 
In \modelname, we delegate the non-private data related computations to server. One would be curious about what if these computations are also performed by data holders using existing secure neural network protocols, i.e., SecureML \cite{mohassel2017secureml}. 
To answer this question, we compare \modelname~with the secure GNN model that is implemented using SecureML, which we call as SecureGNN, where we use 3-degree Taylor expansion to approximate TanH and Sigmoid. 
The accuracy and running time per epoch (in seconds) of \modelname~vs. SecureGNN on Pubmed are 0.8090 vs. 0.7970 and 18.65 vs. 166.81, respectively, where we use local area network. 

\nosection{Analysis of Result3} 
From the above comparison results, we find that our proposed \modelname~learning paradigm not only achieves better accuracy, but also has much better efficiency. 
This is because the non-private data related computations involve many non-linear functions that are not cryptographically friendly, which have to be approximately calculated using time-consuming MPC techniques in SecureML. 

\begin{table}[t]
\centering
\small
\begin{tabular}{cccc}
  \toprule
  Model & \modelname\_C & \modelname\_M & \modelname\_R \\
  \midrule
  Prop.=9:1 & 0.809 & 0.805 & 0.809 \\
  Prop.=8:2 & 0.802 & 0.796 & 0.807 \\
  Prop.=7:3 & 0.793 & 0.793 & 0.803 \\
  \bottomrule
\end{tabular}
\vskip -0.1in
\caption{Comparison of combination operators on Cora by varying the proportion of data hold by $\mathcal{A}$ and $\mathcal{B}$ (Q4).}
\label{regression}
\vskip -0.1in
\end{table}

\begin{table}[t]
\centering
\small
\begin{tabular}{cccc}
  \toprule
  No. of DH & \modelname\_C & \modelname\_M & \modelname\_R \\
  \midrule
  2 & 0.790 & 0.809 & 0.802 \\
  3 & 0.749 & 0.774 & 0.760 \\
  4 & 0.712 & 0.733 & 0.722 \\
  \bottomrule
\end{tabular}
\vskip -0.1in
\caption{Comparison results on Cora by varying the number of data holders (Q5).}\label{ablation-dh}
\vskip -0.15in
\end{table}

\subsection{Ablation Study}
We now study the effects of different combination operators and different number of data holders on \modelname. 

\nosection{Result4: answer to Q4} 
Different combination operators can utilize local node embeddings in diverse ways and make our proposed \modelname~adaptable to different scenarios, we study this by varying the proportion (Prop.) of data (node features and edges) hold by $\mathcal{A}$ and $\mathcal{B}$ in $\{9\colon1, 8\colon2, 7\colon3\}$. 
The results on Cora dataset are shown in Table \ref{regression}. 

\nosection{Analysis of Result4} 
From Table \ref{regression}, we find that with the proportion of data hold by $\mathcal{A}$ and $\mathcal{B}$ being even, i.e., from $\{9\colon1\}$ to $\{7\colon3\}$, the performances of most strategies tend to decrease. 
This is because the neighbor aggregation is done by data holders individually, and with a bigger proportion of data hold by a single holder, it is easier for this party to generate better local node embeddings. 
Moreover, we also find that Mean operator works well when data are evenly split, and Regression operator is good at handling the situations where data holders have different quatity of data, since it treats the local node embeddings from each data holder differently, and assigns weights to them intelligently. 


\nosection{Result5: answer to Q5} 
We vary the number of data holders in \{2, 3, 4\} and study the performance of \modelname. 
We report the results in Table \ref{ablation-dh}, where we use the Cora dataset and assume data holders have even feature and edge data. 

\nosection{Analysis of Result5} 
From Table \ref{ablation-dh}, we find that, as the number of data holders increases, the accuracy of all the models decreases. 
This is because the neighborhood aggregation in \modelname~is done by each holder individually for privacy concern, and each data holder will have less edge data when there are more data holders, since they split the original edge information evenly. 
Therefore, when more participants are involved, more information will be lost during the neighborhood aggregation procedure. 


\begin{table}[t]
\centering
\small
\begin{tabular}{cccccc}
  \toprule
  $\epsilon$ & 4 & 8 & 16 & 32 & 64\\
  \midrule
  Gaussian & 0.502 & 0.702 & 0.772 & 0.789 & 0.794\\
  James-Stein & 0.510 & 0.706 & 0.781 & 0.799 & 0.804\\
  \bottomrule
\end{tabular}
\vskip -0.1in
\caption{Effect of DP on \modelname~using Cora dataset (Q6).}\label{fig:dp}
\vskip -0.15in
\end{table}

\nosection{Result6: answer to Q6} 
We present the privacy loss of each iteration in Table~\ref{fig:dp} and the over all privacy in Theorem~\ref{thm:privacy_all}. %
We vary $\epsilon$ and set $\delta=1e^{-4}$ to study the effects of DP on \modelname. We report the results in Table \ref{fig:dp}, where we use Cora dataset, use MEAN as the combination operator, and assume data holders have even feature and edge data. 

\nosection{Analysis of Result6} 
From Table \ref{fig:dp}, we can see that the accuracy of \modelname~increases with $\epsilon$. In other words, there is a trade-off between accuracy and privacy. The smaller $\epsilon$, the more noise will be added into the local node embeddings, which causes stronger privacy guarantee but lower accuracy. 
We also find James-Stein estimator consistently works better than Gaussian mechanism, since it can reduce MSE, as we have analyzed in Section \ref{sec:model:dp}. 

\section{Conclusion}\label{conclusion}
We propose \modelname, a vertically federated GNN learning paradigm for privacy-preserving node classification task. 
We finish this by splitting the computation graph of GNN. 
We leave the private data related computations on data holders and delegate the rest computations to a server.
Experiments on real world datasets demonstrate that our model significantly outperforms the GNNs by using the isolated data and has comparable performance with the traditional GNN by using the mixed plaintext data insecurely. 

\section*{Acknowledgements}
This work was supported in part by the National Natural Science Foundation of China (No. 62172362) and ``Leading Goose'' R\&D Program of Zhejiang (No. 2022C01126).

\bibliographystyle{named}
\bibliography{ppgnn}

\appendix

\section{Multiplication of Secret Sharing}
Existing research on secret sharing multiplication protocol is mainly based on Beaver's triplet technique \cite{beaver1991efficient}. Specifically, to multiply two secretly shared values $\langle a \rangle$ and $\langle b \rangle$ between two parties $\mathcal{P}_0$ and $\mathcal{P}_1$, they first need to collaboratively choose a shared triple $\langle u \rangle$, $\langle w \rangle$, and $\langle z \rangle$, where $u, w$ are uniformly random values in $\mathds{Z}_{2^\ell}$ and $z=uw$ mod $2^\ell$. 
They then locally compute $\langle e \rangle _i = \langle a \rangle _i - \langle u \rangle _i$ and $\langle f \rangle _i = \langle b \rangle _i - \langle w \rangle _i$, where $i \in \{0, 1\}$. 
Next, they run $\textbf{Rec}(\langle e \rangle _0, \langle e \rangle _1)$ and $\textbf{Rec}(\langle f \rangle _0, \langle f \rangle _1)$. 
Finally, $\mathcal{P}_i$ gets $\langle c \rangle _i = -i \cdot e \cdot f + f \cdot \langle a \rangle _i + e \cdot \langle b \rangle _i + \langle z \rangle _i$ as a share of the multiplication result, such that $\langle a \rangle \langle b \rangle =\langle c \rangle_0+\langle c \rangle_1$. 
It is easy to vectorize the addition and multiplication protocols under secret sharing setting. 
The above protocols work in finite field, and we adopt fixed-point representation to approximate decimal arithmetics efficiently \cite{mohassel2017secureml}.

\section{Related Work in Details}\label{background}
We review three kinds of existing Privacy-Preserving Neural Network (PPNN) models, including GNN models. 

\nosection{PPNN based on cryptographic methods}
These methods mainly use cryptographic techniques, e.g., secret sharing and homomorphic encryption, to build approximated neural networks models \cite{mohassel2017secureml,wagh2019securenn}, since the nonlinear active functions are not cryptographically computable. 
Cryptograph based neural network models are difficult to scale to deep networks and large datasets due to its high communication and computation complexities. 
In this paper, we use cryptographic techniques for data holders to calculate the initial node embeddings securely. 


\nosection{PPNN based on federated learning}
The privacy issues in machine learning has boosted the development of federated learning. To date, federated neural networks have been extensively studied \cite{konevcny2016federated,bonawitz2019towards} and applied into real-world scenarios \cite{li2020review}. 
There are also several work on federated GNN when data are horizontally partitioned \cite{zheng2021asfgnn,wu2021fedgnn}. 
However, a mature solution for federated GNN models under vertically partitioned data is still missing. In this paper, we fill this gap by proposing \modelname.

\nosection{PPNN based on split computation graph} 
These methods split the computation graph of neural networks into two parts, and let data holders calculate the private data related computations individually and get a hidden layer, and then let a server makes the rest computations \cite{vepakomma2018split,chi2018privacy,osia2019hybrid,gu2018securing,zheng2020industrial}. Our model differs from them in mainly two aspects. First, we train a GNN rather than a neural network. Seconds, we use cryptographic techniques for data holders to calculate the initial node embeddings collaboratively rather than compute them based on their plaintext data individually. 

\begin{algorithm}[t]
\caption{Data holders $\mathcal{P}$ securely generate the initial node embeddings using secret sharing}\label{algo-horLR}
\textbf{Input}: \{$\textbf{x}^i_v \in \mathds{Z}_{2^\ell}, \forall v \in \mathcal{V}, \forall i \in \mathcal{P}$\}, and $\textbf{x}^i$ for short \\
\textbf{Output}: The share of initial node embeddings for each data holder $i$ \{$\textbf{h}_v^0(i), i \in \mathcal{P}, \forall v \in \mathcal{V}$\} 
\begin{algorithmic}[1]
\FOR{$\mathcal{P}_i \in \mathcal{P}$ \texttt{in parallel}}
	\STATE $\mathcal{P}_i$ randomly initializes $\langle \textbf{W} \rangle _i $
	\STATE $\mathcal{P}_i$ locally generates $\left\{\left\langle\textbf{x}^{i}\right\rangle_j\right\}_{j\in \mathcal{P}}$ \label{algo-init-ss1}
	\STATE $\mathcal{P}_i$ distributes $\left\{\left\langle\textbf{x}^{i}\right\rangle_j\right\}_{j\neq i}$ to others \label{algo-init-ss2}
	\STATE $\mathcal{P}_i$ concats $\left\{\left\langle\textbf{x}^{k}\right\rangle_j\right\}_{j \in \mathcal{P}}$ and get $\left\langle\textbf{x}\right\rangle_i $ \label{secredemb-order} 
	\STATE $\mathcal{P}_i$ locally calculates $\left\langle\textbf{x}\right\rangle_i ^T \cdot \langle \textbf{W} \rangle _i$ as $i$-share \label{algo-init-smm1}
\ENDFOR

\FOR{$\mathcal{P}_j \in \mathcal{P}$ and $j \neq i$}
	\STATE $\mathcal{P}_i$ and $\mathcal{P}_j$ calculate $i$-share $\left\langle\left\langle\textbf{x}\right\rangle_i ^T \cdot \langle \textbf{W} \rangle _j\right\rangle_i$ and $j$-share $\left\langle\left\langle\textbf{x}\right\rangle_i ^T \cdot \langle \textbf{W} \rangle _j\right\rangle_j$ using secret sharing \label{algo-init-smm2} 
	\STATE $\mathcal{P}_i$ and $\mathcal{P}_j$ calculate $i$-share $\left\langle\left\langle\textbf{x}\right\rangle_j ^T \cdot \langle \textbf{W} \rangle _i\right\rangle_i$ and $j$-share $\left\langle\left\langle\textbf{x}\right\rangle_j ^T \cdot \langle \textbf{W} \rangle _i\right\rangle_j$ using secret sharing \label{algo-init-smm3} 
\ENDFOR

\FOR{$\mathcal{P}_i \in \mathcal{P}$ \texttt{in parallel}}
    \STATE $\mathcal{P}_i$ locally calculates the summation of all $i$-shares, denoted as $\left\langle\textbf{h}_v^0\right\rangle_i = \left\langle\textbf{x} ^T \cdot \textbf{w}\right\rangle_i$ 
    \STATE $\mathcal{P}_i$ sends $\left\langle\textbf{h}_v^0\right\rangle_i$ to $\mathcal{P}_j$, $\forall j \in \mathcal{P}, j \neq i$ 
    \STATE $\mathcal{P}_i$ reconstructs $\textbf{h}_v^0$, denote as $\textbf{h}_v^0(i)$
\ENDFOR

\STATE \textbf{return} $\textbf{h}_v^0(i)$ for each data holder $i \in \mathcal{P}$
\end{algorithmic}
\end{algorithm}

\section{Algorithm of Generating Initial Node Embeddings}\label{sec-model-aggre}

We summarize the algorithm of generating initial node embeddings in Algorithm 1. 
Traditionally, the initial node embeddings can be generated by $\textbf{h}_0=\textbf{x} ^T \cdot \textbf{W}$, where $\textbf{x}$ is node features and $\textbf{W}$ is weight matrix. 
When features are vertically partitioned, we calculate initial node embeddings as follows. 
First, we secretly share $\textbf{x}$ among data holders (Lines 3-4). 
Then, data holders concat their received shares in order (Line 5). 
After it, we calculate $\textbf{x} ^T\cdot\textbf{W}$ following distributive law (Lines 8-11). 
Take two data holders for example,  $\textbf{x} ^T \cdot \textbf{W} = (\left\langle\textbf{x}\right\rangle_1+\left\langle\textbf{x}\right\rangle_2) \cdot (\left\langle\textbf{W}\right\rangle_1+\left\langle\textbf{W}\right\rangle_2)=
\left\langle\textbf{x}\right\rangle_1 \cdot \left\langle\textbf{W}\right\rangle_1 $ $+\left\langle\textbf{x}\right\rangle_1 \cdot \left\langle\textbf{W}\right\rangle_2+\left\langle\textbf{x}\right\rangle_2 \cdot \left\langle\textbf{W}\right\rangle_1+\left\langle\textbf{x}\right\rangle_2 \cdot \left\langle\textbf{W}\right\rangle_2$. 
Finally, data holders reconstruct $\textbf{x} ^T \cdot \textbf{W}$ by summing over all the shares (Lines 12-16).

\end{document}